\documentclass[conference]{IEEEtran}
\IEEEoverridecommandlockouts
\usepackage{cite}
\usepackage{amsmath,amssymb,amsfonts}
\usepackage{algorithmic}
\usepackage{graphicx}
\usepackage{textcomp}
\usepackage{booktabs}
\usepackage{multirow}
\usepackage{makecell}
\usepackage{url}
\usepackage{array}
\usepackage{cleveref}
\usepackage{amsmath}
\crefname{subsection}{Appendix}{Appendices}
\usepackage{xcolor}
\usepackage{soul}
\usepackage[table]{xcolor}
\def\BibTeX{{\rm B\kern-.05em{\sc i\kern-.025em b}\kern-.08em
    T\kern-.1667em\lower.7ex\hbox{E}\kern-.125emX}}

\begin{document}

\title{\textit{Multiverse}: Language-Conditioned Multi-Game Level Blending via Shared Representation \\
}

\author{
\IEEEauthorblockN{
In-Chang Baek\IEEEauthorrefmark{1}\IEEEauthorrefmark{2}\thanks{*These authors contributed equally to this work.},
Jiyun Jung\IEEEauthorrefmark{1}\IEEEauthorrefmark{3},
Geum-Hwan Hwang\IEEEauthorrefmark{2},
Sung-Hyun Kim\IEEEauthorrefmark{2},
Kyung-Joong Kim\IEEEauthorrefmark{2}\IEEEauthorrefmark{4}\thanks{\S{}Corresponding author.}
}

\IEEEauthorblockA{\IEEEauthorrefmark{2}Gwangju Institute of Science and Technology (GIST), South Korea}
\IEEEauthorblockA{\IEEEauthorrefmark{3}Dongguk University, South Korea}

\IEEEauthorblockA{
inchang.baek@gm.gist.ac.kr
}
}
\maketitle

\newcommand{\jy}[1]{\textcolor{purple}{#1}}
\newcommand{\ic}[1]{\textcolor{blue}{#1}}
\newcommand{\gh}[1]{\textcolor{orange}{#1}}

\newcommand{\tbd}[1]{\textcolor{red}{#1}}

\newcommand{\hlbox}[2]{%
  {\setlength{\fboxsep}{0pt}\colorbox{#1}{#2}}%
}
\newcommand{\hlA}[1]{\hlbox{green!20}{#1}}
\newcommand{\hlB}[1]{\hlbox{red!18}{#1}}

\newcommand{\graycell}[0]{\cellcolor{gray!15}}
\newcommand{\skycell}[0]{\cellcolor{blue!5}}

\newcommand{\up}[1]{\textcolor{blue!60!black}{#1}}
\newcommand{\down}[1]{\textcolor{red!60!black}{#1}}

\begin{abstract}
Text-to-level generation aims to translate natural language descriptions into structured game levels, enabling intuitive control over procedural content generation.
While prior text-to-level generators are typically limited to a single game domain, extending language-conditioned generation to multiple games requires learning representations that capture structural relationships across domains.
We propose \textbf{Multiverse}, a language-conditioned multi-game level generator that enables cross-game level blending through textual specifications.
The model learns a shared latent space aligning textual instructions and level structures, while a threshold-based multi-positive contrastive supervision links semantically related levels across games.
This representation allows language to guide which structural characteristics should be preserved when combining content from different games, enabling controllable blending through latent interpolation and zero-shot generation from compositional textual prompts.
Experiments show that the learned representation supports controllable cross-game level blending and significantly improves blending quality within the same game genre, while providing a unified representation for language-conditioned multi-game content generation.
\end{abstract}


\section{Introduction}
Text-to-level generation has emerged as a practical interface for procedural content generation, enabling models to synthesize game levels directly from natural-language specifications. Prior work has shown that language-conditioned generators can translate textual descriptions into structured level layouts~\cite{merino2023five, sudhakaran2023mariogpt}. More recent studies increasingly frame this problem as representation learning, where models learn text–level aligned embeddings either by encoding language conditions for downstream optimization or by aligning semantic descriptions with level structures through contrastive objectives~\cite{baek2025ipcgrl, baek2025human}. While these approaches successfully establish language–level alignment, most are developed within a single game domain, limiting their ability to capture structural correspondences across heterogeneous games. This limitation motivates learning shared level representations that explicitly connect structures across multiple game domains.

\begin{figure}[!t]
    \centering
    \includegraphics[width=0.9\linewidth]{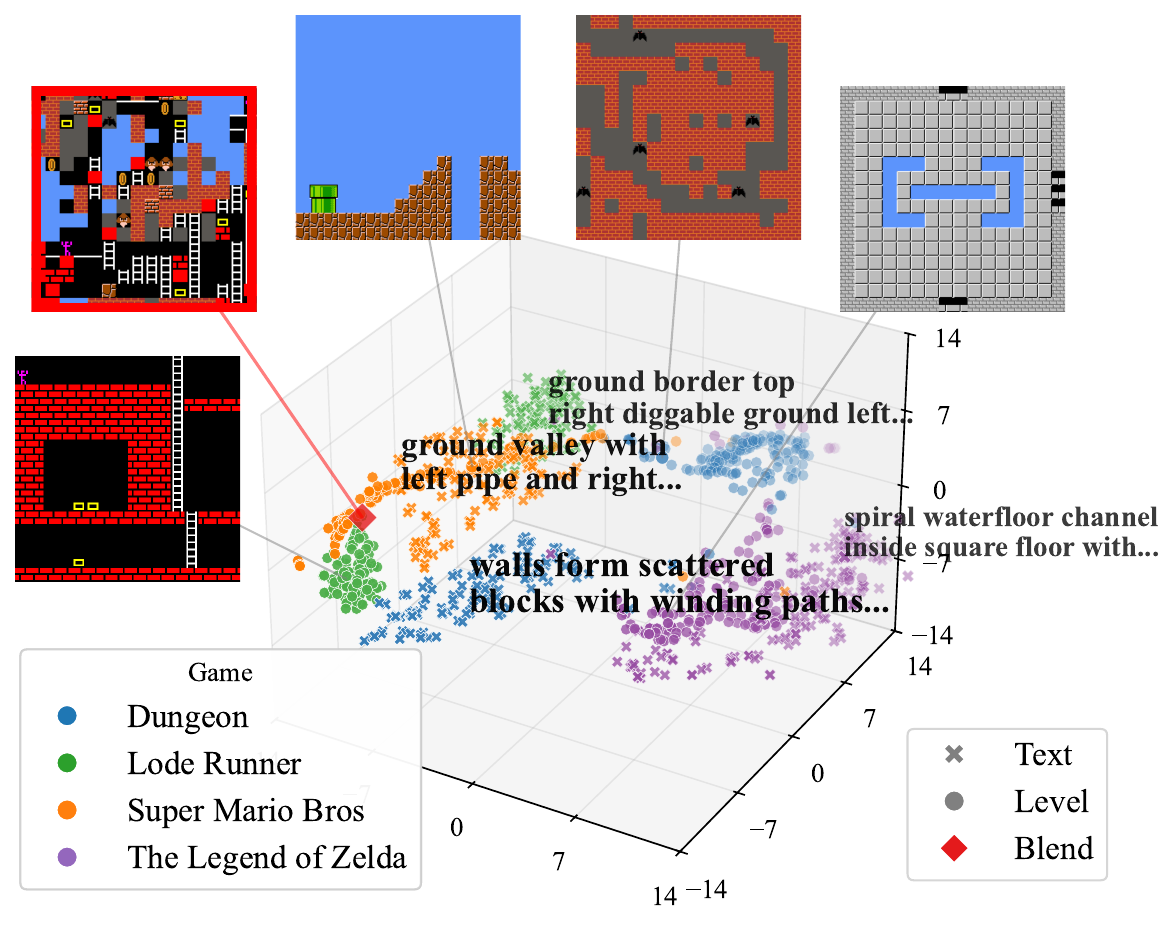}
    \vspace{-0.3cm}
    \caption{\textbf{Shared Latent Space Visualization.}
    The proposed inter-game contrastive learning learns 128-dimensional representations for levels and instructions across multiple games, which are projected into three dimensions using t-SNE.
    Embeddings from different games are aligned within a unified latent space. The blended sample is generated by an interpolation between \emph{Dungeon} and \emph{Lode Runner} level embeddings, demonstrating cross-game continuity in the shared space.
    }
    \label{fig:shared_rep}
    \vspace{-0.5cm}
\end{figure}

A complementary line of work explores cross-game relationships through level blending, where levels from different games are connected within a shared representation space. Prior approaches combine domain features through latent interpolation or conditional generation (e.g., VAE-based controllable blending~\cite{sarkar2020controllable} and shared latent models for multi-domain level generation~\cite{kumaran2019generating}), while other methods treat the latent space as a design space for exploration, such as tile embedding–based shared feature spaces~\cite{atmakuri2025semi} and quality-diversity search over generative latent spaces~\cite{fontaine2021illuminating, sarkar2023procedural}. However, existing blending methods rely on latent manipulation rather than natural-language conditioning, while prior text–level representation learning focuses on single-domain alignment and does not model cross-game structural relationships~\cite{baek2025human}.

To fill this research gap, we propose \textit{Multiverse}, a language-conditioned multi-game generator that enables cross-game blending through natural-language specifications. Fig.~\ref{fig:shared_rep} shows the shared latent space learned by Multiverse. Our key idea is to align text and level representations while establishing structural correspondences across game domains, allowing language to steer generation across games. We introduce a \emph{cross-game contrastive learning} formulation that aligns game domains while preserving fine-grained semantic structure across text and level modalities. Experiments on multi-game level generation and cross-game blending demonstrate improved blending quality over a standard contrastive baseline, supported by semantic similarity evaluation of generated levels.

We summarize our main contributions as follows:
\begin{itemize}
  \item A representation learning framework for a shared text–level latent space via cross-game structural connectivity.
  \item The proposed text-level representation space improves controllability and generation quality in inter-game level generation.
  \item Demonstration that the learned representation enables zero-shot cross-game level generation from compositional textual instructions.
\end{itemize}

\section{Background}
\subsection{Text-to-Level Representation Learning}
Text-to-level generation aims to synthesize game levels from natural-language descriptions, translating high-level design intent into structured artifacts such as discrete tilemaps. Early approaches primarily focused on conditioning neural generators on sentence embeddings to align textual descriptions with spatial level structures~\cite{merino2023five}, with subsequent work exploring alternative training paradigms such as distillation-based frameworks for text-conditioned generation~\cite{nie2025moonshine}. In contrast, reinforcement learning–based formulations treat natural-language instructions as objectives, where representation learning enables instruction embeddings to function as conditional features for the generation policy~\cite{baek2025ipcgrl,kim2025multi}. More recent work further explores representation learning through contrastive objectives, using CLIP-based alignment~\cite{radford2021learning} to capture semantic conditions that are difficult to express with explicit numerical features, enabling unsupervised alignment between textual descriptions and game levels~\cite{baek2025human}.

However, extending text-to-level generation to multi-game settings introduces additional challenges, as structurally similar gameplay concepts are often described using different vocabularies across games, disrupting text–level alignment. While prior work establishes such alignment in single-domain settings, transferring it across games remains challenging. This motivates learning shared text–level representations that capture cross-game relationships. In this work, we leverage structural connectivity between levels to strengthen multi-game alignment while learning a generator-agnostic representation that can be integrated with diverse generation frameworks.
These challenges become particularly important when extending text-conditioned generation to cross-domain level blending.

\begin{figure*}[!ht]
    \centering
    \includegraphics[width=1\linewidth]{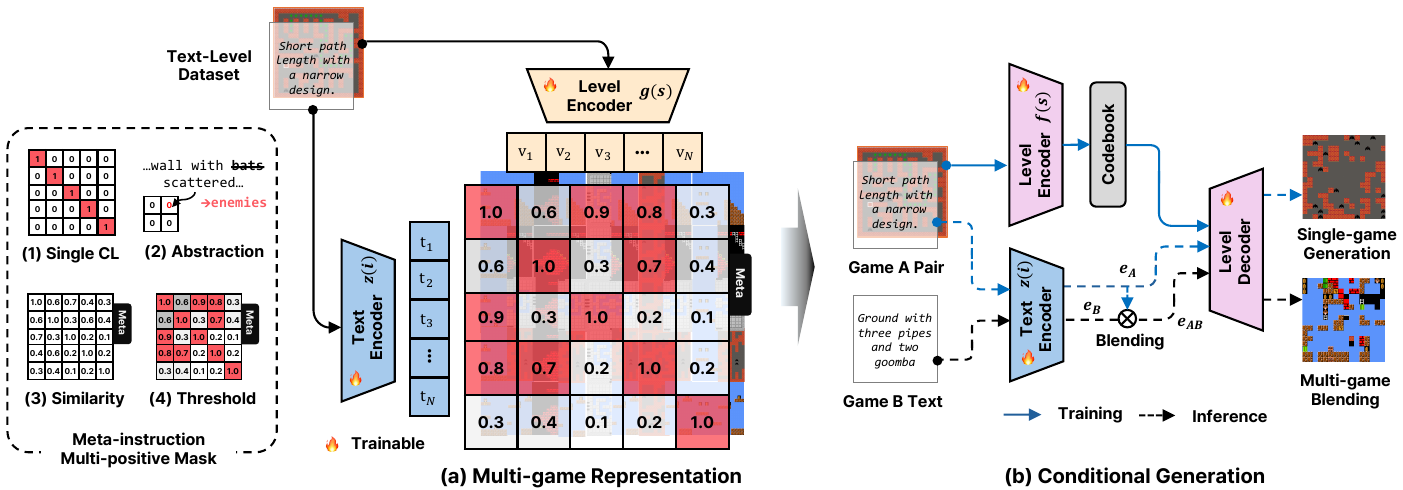}
    \vspace{-0.6cm}
    \caption{\textbf{\textit{Multiverse} Training Pipeline and Level Blending Mechanism.} (a) A shared multi-game latent space is learned using multi-positive contrastive learning over text--level pairs. (b) The learned level embedding is used as a conditional vector for a VAE-based generator. For single-game generation, the encoded embedding directly conditions the decoder. For multi-game level blending, embeddings from different games are interpolated in the latent space, and the interpolated vector is provided to the decoder to synthesize blended levels.
    }
    \label{fig:architecture}
    \vspace{-0.3cm}
\end{figure*}

\subsection{Level Blending}
Level blending synthesizes novel levels by combining structural attributes from multiple source levels or game domains. 
A common formulation learns a continuous latent representation of levels and produces blends through interpolation or conditional sampling in that space. 
For instance, conditional generative models such as CVAEs enable controllable blending by introducing conditioning variables that encode structural properties or connectivity constraints of level segments~\cite{sarkar2020controllable, sarkar2021dungeon}. 
Other approaches treat the learned latent space as a search domain, where evolutionary or quality-diversity algorithms explore candidate blends while enforcing playability or structural constraints~\cite{fontaine2021illuminating,sarkar2021generating,sarkar2023procedural}. 
Additionally, tile embedding methods learn vector representations of tiles across multiple games, enabling shared feature spaces that support cross-domain level generation and blending~\cite{atmakuri2025semi}.

To support cross-domain generation more broadly, subsequent work has explored multi-domain generative models that learn shared latent spaces across games~\cite{kumaran2019generating}. 
However, these pipelines typically rely on latent-space manipulation or non-linguistic control variables rather than natural-language specifications of design intent. 
In contrast, we propose a text-guided multi-game learning approach that aligns language with a shared level representation, enabling controllable cross-game blending driven directly by textual descriptions.

\section{Multi-game Level Dataset}
To support language-conditioned multi-game generation, we construct a dataset composed of levels from multiple games spanning two distinct genres: top-down dungeon crawler and platformer. These two genres exhibit distinct structural properties in their level layouts. For the top-down dungeon crawler genre, we use level data from \emph{The Legend of Zelda} obtained from the Video Game Level Corpus (VGLC) \cite{VGLC}, along with a \emph{Dungeon} dataset \cite{baek2025human} from PCGRL environment \cite{khalifa2020pcgrl}. For the platformer genre, we use levels from \emph{Lode Runner} and \emph{Super Mario Bros.} obtained from the VGLC.
In particular, platformer games involve gravity-driven movement and vertical traversal, while top-down dungeon crawlers emphasize planar navigation and room connectivity.

\textbf{Preprocessing.}
To ensure consistency across games, we standardize all level data to a fixed size of $16 \times 16$. The dataset contains a total of 5,576 levels across all games, and each tile is represented using a one-hot encoding over a vocabulary of 34 tile types. Larger levels are divided into smaller sub-levels using a sliding window approach to preserve local spatial structure, while smaller levels are padded by replicating surrounding tiles to maintain boundary continuity. To balance the number of training samples across domains, we apply game-dependent data augmentation. Levels from platformer games are augmented using horizontal flipping, while levels from dungeon crawler games are augmented using both rotation and horizontal flipping to increase structural diversity.

\textbf{Annotation.}
To enable text–level alignment across multiple game domains, we generate language annotations for each level using a vision-language model (VLM). Each preprocessed level, represented as a two-dimensional integer array, is rendered into an image using game-specific tile sprites and annotated by the VLM to produce textual descriptions capturing structural properties such as object distribution and spatial layout. To obtain consistent annotations across games, we design a structured prompting scheme that constrains the vocabulary to gameplay-relevant entities (e.g., ground, question block, goomba, coin, cannon) and instructs the model to describe object distribution without referring to visual attributes such as color. We use \texttt{gpt-5.2-2025-12-11}\footnote{\url{https://platform.openai.com/docs/models/gpt-5.2}} for this annotation process. 
The resulting descriptions average $11.2\pm{}2.4$ words and serve as instruction annotations.

\section{Language-Conditioned Multi-game Level Generator}

\subsection{Multi-game Shared Representation}
A key challenge in multi-game representation learning is capturing structural commonalities across levels described with different domain-specific vocabularies, which conventional contrastive learning (CL) struggles to model due to its single-positive pairing assumption.
Treating such related instructions as negatives hinders the formation of a coherent shared representation. To address this issue, we introduce a meta-instruction abstraction process with a similarity-based multi-positive CL framework. As illustrated in Fig.~\ref{fig:architecture}, instructions are first abstracted to reduce domain-specific variations, and semantic similarity is used to construct a multi-positive mask for cross-game alignment. The resulting representation then conditions the generator to enable both single-game generation and cross-game level blending.

\textbf{Single-positive CL.} As a baseline, we adopt the standard single-positive CL \cite{radford2021learning}, where each level is paired only with its corresponding instruction, while all other samples in the batch are treated as negatives. This objective does not explicitly encourage semantic grouping across levels or game domains. This formulation follows the widely adopted paradigm of contrastive representation learning for visual–text alignment, and has been adopted in language-conditioned level generation frameworks~\cite{baek2025human}. Let $\mathbf{v}_i \in \mathbb{R}^{d}$ and $\mathbf{t}_i \in \mathbb{R}^{d}$ denote the projected embeddings of the $i$-th level and instruction, respectively, and let $\tau > 0$ denote a temperature parameter. The model is trained using the InfoNCE loss.

\textbf{Instruction abstraction.}
Since different games often describe similar gameplay concepts using different vocabulary, we normalize domain-specific terms to enable semantic comparison across games. Instruction abstraction reduces domain-specific vocabulary differences by replacing game-specific terms with generalized expressions shared across domains. For example, a term such as \textit{bat} is replaced with the generalized term \textit{enemy}. We implement this step using a small set of rule-based lexical substitutions (15 rules) that map domain-specific entities to shared semantic categories such as enemies, environmental elements, climbable structures, collectables, interactive blocks, and hazards. Formally, given an instruction $s$, the abstraction function $A(\cdot)$ produces a generalized instruction $\hat{s} = A(s)$. This transformation removes domain-dependent surface forms while preserving high-level semantics, enabling cross-domain alignment during CL.

\textbf{Similarity-based multi-positive CL.} After abstraction, a pretrained text embedding model $f(\cdot)$ encodes each instruction to estimate semantic similarity. Cosine similarity $\mathrm{sim}\!\left(\cdot, \cdot\right)$ between instruction embeddings defines semantic relationships among instructions. This similarity estimation identifies related instructions even when they originate from different game domains. We use the \texttt{text-embedding-3-large}\footnote{\url{https://platform.openai.com/docs/models/text-embedding-3-large}} model as the text embedding model, which produces 3072-dimensional embedding vectors for input text.
Based on the similarity scores, a threshold $\delta$ determines the final positive mask $\mathcal{P}_{\mathrm{meta}}(\cdot)$ for CL. Levels whose generalized instruction similarity exceeds $\delta$ are treated as additional positive pairs, as defined in Eq.~(\ref{eq:meta_positive}). The final contrastive objective follows the InfoNCE\cite{radford2021learning} formulation, averaging the level-to-text and text-to-level losses as shown in Eq.~(\ref{eq:meta_loss}). This strategy promotes cross-game semantic alignment by ignoring domain-specific surface forms and encouraging levels with similar high-level structures to occupy nearby regions in the shared latent space.
\begin{equation}
\mathcal{P}_{\mathrm{meta}}(i)
=
\left\{ j \mid \mathrm{sim}\!\left(f(\hat{s}_i), f(\hat{s}_j)\right) > \delta,\; j \neq i \right\}
\label{eq:meta_positive}
\end{equation}
\begin{align}
\mathcal{L}_{\mathrm{meta}}
=
-\frac{1}{2}
\sum_i
\Bigg[
&
\log
\frac{
\sum\limits_{j \in \mathcal{P}_{\mathrm{meta}}(i)}
\exp(\mathbf{v}_i^{\top}\mathbf{t}_j / \tau)
}{
\sum\limits_{k}
\exp(\mathbf{v}_i^{\top}\mathbf{t}_k / \tau)
}
\nonumber \\
&+
\log
\frac{
\sum\limits_{j \in \mathcal{P}_{\mathrm{meta}}(i)}
\exp(\mathbf{v}_j^{\top}\mathbf{t}_i / \tau)
}{
\sum\limits_{k}
\exp(\mathbf{v}_k^{\top}\mathbf{t}_i / \tau)
}
\Bigg]
\label{eq:meta_loss}
\end{align}

\subsection{Shared-Latent Level Generator}
We implement a conditional VQ-VAE~\cite{van2017neural} as our level generator, where the shared embedding space steers the decoding process.
Let $x$ denote a game level represented as a tile-wise tensor $x \in \mathbb{R}^{H \times W \times C}$, where $H{=}W{=}16$ and $C$ denotes the number of tile types, with each spatial location encoded as a one-hot vector over the $C$ tile categories.
The encoder maps $x$ to a spatial latent grid $\mathbf{z}_e \in \mathbb{R}^{D \times 4 \times 4}$, where $D$ denotes the latent embedding dimension and the $4 \times 4$ grid corresponds to the downsampled spatial resolution of the input level. The latent vectors are then discretized via nearest-neighbor lookup in a codebook of size $K$ to obtain the quantized latent $\mathbf{z}_q$.
We adopt the exponential moving averages (EMA)-based vector quantization scheme of Kaiser et al.~\cite{kaiser2018fast}, where codebook entries are updated using exponential moving averages of encoder assignments rather than gradient-based optimization.
This strategy stabilizes codebook learning and mitigates code collapse during training.

For a sampled text–level pair $(s,x)$, the instruction $s$ is encoded into a text embedding $\mathbf{t}$ using the text encoder.
The resulting embedding is projected by $g(\cdot)$ and concatenated with the quantized latent to form the conditioned latent representation $\mathbf{z}_{\mathrm{cond}} = [\mathbf{z}_q ; g(\mathbf{t})]$, where $[\cdot ; \cdot]$ denotes channel-wise concatenation.
The decoder then reconstructs the level as $\hat{x} = D(\mathbf{z}_{\mathrm{cond}})$.
The generator is trained with the VQ-VAE objective defined in Eq.~\eqref{eq:generator_loss}:

\begin{equation}
\mathcal{L}_{\mathrm{generator}}
=
\mathcal{L}_{\mathrm{BCE}}(x,\hat{x})
+
\beta \left\| \mathrm{sg}[\mathbf{z}_q] - \mathbf{z}_e \right\|_2^2,
\label{eq:generator_loss}
\end{equation}

where $\mathcal{L}_{\mathrm{BCE}}$ denotes the binary cross-entropy reconstruction loss between the ground-truth level $x$ and the reconstructed level $\hat{x}$, $\mathrm{sg}[\cdot]$ denotes the stop-gradient operator, and $\beta$ controls the commitment to discrete codes. During training, the projection network $g(\cdot)$ is optimized jointly with the generator, enabling gradients from $\mathcal{L}_{\mathrm{generator}}$ to flow through the conditioning pathway and align the projected text embedding with the generator's discrete latent space.

\section{Experimental Setup} 

\subsection{Research Question}
\textbf{Single-game instruction.} 
\textit{Can a unified multi-game generator maintain competitive performance compared to individually trained single-game models?}
We condition the generator on original single-game instructions and generate levels for each domain. 
The generated levels are evaluated by measuring visual similarity to reference levels from the corresponding game, assessing whether the shared latent representation preserves in-domain generation quality despite multi-game training.

\textbf{Embedding interpolation.} 
\textit{Does the proposed representation learning enable controllable cross-game level blending through interpolation in the shared embedding space?}
Level pairs from two games are sampled and encoded into the shared embedding space. 
Interpolated embeddings with varying ratios (e.g., 50:50) are used to generate levels. 
We then measure how visual similarity shifts toward each source domain as the interpolation ratio changes. 
We evaluate whether interpolation in the shared embedding space produces structurally coherent blended levels, extending tile-based blending approaches to the level structure domain~\cite{atmakuri2025semi}.

\textbf{Multi-game instruction.} \textit{Can composite instructions combining two game domains enable zero-shot blended level generation, and does instruction composition control the relative contribution of each domain in the generated levels?}
We compare four instruction composition strategies that represent different ways of combining two game-domain instructions. Generated levels are evaluated by visual similarity to the reference levels associated with each of the two original instructions, and the composite instruction embedding is analyzed relative to the two source instruction embeddings in the shared space.

\subsection{Comparison}
To evaluate the effectiveness of the proposed cross-game contrastive learning, we compare several contrastive supervision strategies under a unified experimental setting. As a reference baseline, we adopt single-positive contrastive learning (\textit{Single CL}) \cite{radford2021learning}, which is commonly used in prior text-to-level generation and contrastive representation learning studies focusing on a single game domain. This baseline allows us to assess whether additional cross-game supervision improves representation quality.

We further conduct an ablation study to analyze the contributions of the proposed meta-instruction framework. The \textit{-Abstraction} variant removes the meta-instruction component, constructing contrastive supervision without meta-instructions. The \textit{-Threshold} variant uses meta-instructions but applies single-positive CL without similarity-based positive mask construction.

\subsection{Metric}
\textbf{ViTScore ($\uparrow$)} \cite{zhu2024evaluate} measures the visual semantic similarity between a generated level and its ground-truth reference using embeddings from a pre-trained vision model.
We evaluate generated levels from a \emph{visual semantic pattern} perspective by rendering each level into an image and extracting embeddings with a pre-trained Vision Transformer (ViT) \cite{dosovitskiy2020vit}.
ViT-based embeddings have been used in game-level domains as semantic signals: they were leveraged to classify specific level shapes \cite{abdullah2024chatgpt4pcg}, and to compare style similarity between generated and human-designed levels \cite{baek2025human}.
In our work, we adopt \textit{ViTScore} to measure whether a generated level is visually and semantically similar to its ground-truth counterpart.

Given a rendered ground-truth level image $x$ and its corresponding generated level image $\hat{x}$, both represented as RGB images, we compute the cosine similarity between their embeddings extracted by a pre-trained ViT\footnote{\url{https://huggingface.co/google/vit-base-patch16-224}} encoder $f_{\mathrm{ViT}}$. The ViTScore is defined as in Eq.~\eqref{eq:vitscore}.

\begin{equation}
\label{eq:vitscore}
\mathrm{ViTScore}(x, \hat{x})
= \frac{f_{\mathrm{ViT}}(x)^\top f_{\mathrm{ViT}}(\hat{x})}{\left\lVert f_{\mathrm{ViT}}(x)\right\rVert_2 \left\lVert f_{\mathrm{ViT}}(\hat{x})\right\rVert_2}.
\end{equation}

This score is computed on the held-out test split (20\% of the dataset, unseen during training), and we report the mean \textit{ViTScore} over each test subset.
Since the similarity range differs across games, we additionally report normalized scores for each game.

\textbf{Weighted Geometric Mean (WGM, $\uparrow$).}
WGM aggregates similarity scores to multiple source domains into a single ratio-aware metric for blended level evaluation.
It is introduced as a quantitative metric for our \textit{embedding interpolation} experiment.
Specifically, let $S_A$ and $S_B$ denote the ViTScore similarity between a generated level and the ground-truth references of domains $A$ and $B$, respectively.
Given a blending ratio $\alpha \in [0,1]$, WGM is defined as in Eq. \eqref{eq:wgm}.

\begin{equation}
\label{eq:wgm}
\mathrm{WGM}_\alpha
= S_A^{\alpha} S_B^{1-\alpha}.
\end{equation}

This formulation follows the standard definition of a weighted geometric mean and evaluates generation quality by jointly considering the ViTScore similarity and the relative importance of each domain defined by the blending ratio.
In contrast to an arithmetic average, which may reward levels that predominantly resemble a single source game, the geometric mean down-weights such one-sided matches and instead favors blended levels that preserve meaningful structural patterns from both domains.

\subsection{Implementation}
\textbf{Representation learning.} 
For the contrastive objective, we use an embedding dimension of 128. The level encoder is a CNN-based residual map encoder followed by global average pooling and a projection head. The text encoder consists of a pretrained CLIP ViT-B/32\footnote{\url{https://huggingface.co/openai/clip-vit-base-patch32}} backbone with a linear projection layer to map text features into the shared embedding space, while keeping the backbone frozen. The contrastive objective uses a learnable temperature parameter initialized at 0.14. A semantic similarity threshold $\delta$ of 0.3 filters inconsistent text–level pairs during training, and the model is optimized with Adam using a learning rate of $5 \times 10^{-5}$.

\textbf{Generator training.} 
For the EMA-VQ-VAE generator, we use a latent embedding dimension $D=128$ and a codebook size $K=256$. The commitment loss weight is initialized to $\beta=0.5$ and is lower-bounded by $\beta_{\min}=0.3$. To stabilize early training, we apply a simple schedule to the commitment term using a multiplicative coefficient that starts at 1.0 and decays to a minimum of 0.6, with the decay starting after 100 steps. Unless otherwise specified, all remaining hyperparameters follow our default training configuration.
The VQ-VAE encoder and decoder are trained with a learning rate of $4\times10^{-4}$.

\textbf{End-to-end training.}
Results are reported as the mean and standard deviation across 10 independent runs with different random seeds.
Experiments are conducted on a single NVIDIA RTX 3090 GPU with a batch size of 512, 2K epochs, and each run takes approximately 1.5 wall-clock hours.
The source code, dataset, and video materials are released for reproducibility\footnote{\url{https://github.com/bic4907/Multiverse-multigame-pcg}}.

\section{Experimental Result}
\subsection{Single-Game Instruction}

\begin{table}[!h]
\centering
\small
\caption{Single-game generation VITScore performance under single-game instructions.}
\label{tab:single_game}
\begin{tabular}{lc|cc}
\toprule
Dataset & Single-game & \multicolumn{2}{c}{Multi-game} \\
\textbf{Game} & \textbf{Single CL} & \textbf{Single CL} & \textbf{Multiverse} \\
\midrule
Dungeon (G1)     & 0.794{\scriptsize$\pm$0.004} 
                 & \textbf{0.730}{\scriptsize$\pm$0.003} 
                 & 0.728{\scriptsize$\pm$0.005} \\

Zelda (G2)       & 0.767{\scriptsize$\pm$0.009} 
                 & 0.687{\scriptsize$\pm$0.016} 
                 & \textbf{0.719}{\scriptsize$\pm$0.020} \\

SuperMario (G3)  & 0.623{\scriptsize$\pm$0.010} 
                 & 0.593{\scriptsize$\pm$0.008} 
                 & \textbf{0.599}{\scriptsize$\pm$0.007} \\

Lode Runner (G4) & 0.792{\scriptsize$\pm$0.005} 
                 & \textbf{0.800}{\scriptsize$\pm$0.003} 
                 & 0.799{\scriptsize$\pm$0.005} \\

\midrule
Overall & 0.744{\scriptsize$\pm$0.082} 
                 & 0.703{\scriptsize$\pm$0.087} 
                 & \textbf{0.711}{\scriptsize$\pm$0.083} \\

\bottomrule
\end{tabular}
\end{table}

Table~\ref{tab:single_game} reports the per-game generation performance under single-game instructions. Compared to individually trained single-game models, Multiverse exhibits only a modest performance drop of about 4.4\% in the overall score, indicating that a single shared embedding space and generator can largely preserve in-domain generation quality despite being trained across multiple domains. Moreover, Multiverse achieves performance comparable to the single-positive baseline across games while slightly improving the overall score, suggesting that the proposed multi-positive alignment preserves single-game fidelity while enabling cross-domain generalization.

\begin{figure*}[!t]
\centering

\setlength{\tabcolsep}{3pt}  
\renewcommand{\arraystretch}{1.0}

\begin{tabular}{ccccccc}
\toprule
Level A & Level B & $\alpha=1.0$ & $\alpha=0.75$ & $\alpha=0.5$ & $\alpha=0.25$ & $\alpha=0.0$ \\
\midrule
\includegraphics[width=2.2cm]{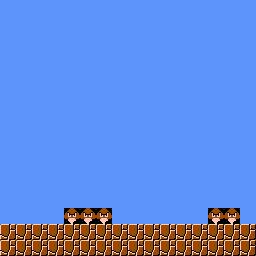} \vspace{-6pt} &
\includegraphics[width=2.2cm]{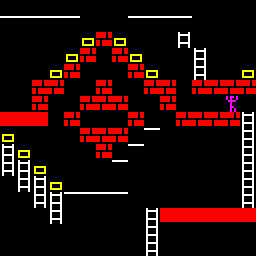} &
\includegraphics[width=2.2cm]{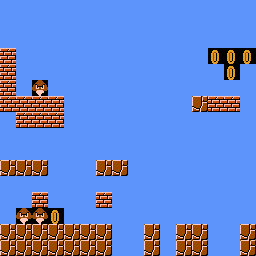} &
\includegraphics[width=2.2cm]{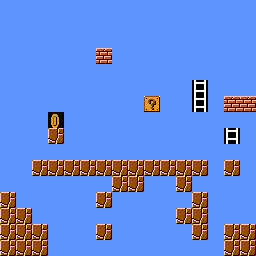} &
\includegraphics[width=2.2cm]{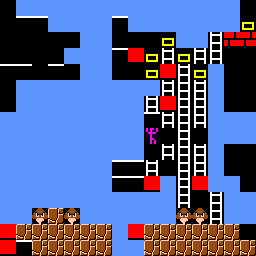}&
\includegraphics[width=2.2cm]{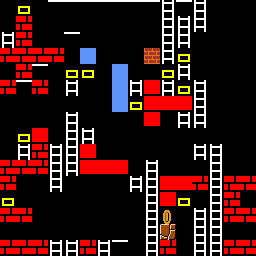}&
\includegraphics[width=2.2cm]{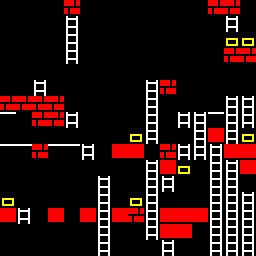} \\[5pt]
\bottomrule 

\multicolumn{7}{p{0.95\linewidth}}{\footnotesize{
A: Flat ground with two small breakable block clusters above \newline
B: Ground bottom right diggable ground center top gold left pink enemy right ladder
}
}

\end{tabular}

\caption{\textbf{Embedding Interpolation-based Level Blending.}
Levels generated by interpolating between the embeddings of Super Mario (A) and Lode Runner (B) across different mixing ratios.}
\vspace{-0.5cm}
\label{fig:embedding_blending}
\end{figure*}

\subsection{Level Blending via Embedding Interpolation}
This experiment investigates whether the shared embedding space enables controllable level blending between two game domains by interpolating between their corresponding text embeddings while varying the blending ratio. We generate levels from interpolated embeddings at ratios $(0{:}100, 25{:}75, 50{:}50, 75{:}25, 100{:}0)$.
Let $\mathbf{t}_A$ and $\mathbf{t}_B$ denote the text embeddings corresponding to instructions from game domains $A$ and $B$, respectively. 
We compute an interpolated embedding using a weighted combination as in Eq.~\eqref{eq:embedding_interp}.

\begin{equation}
\mathbf{t}_{\alpha} = \alpha \mathbf{t}_A + (1-\alpha) \mathbf{t}_B,\quad \alpha \in [0,1]
\label{eq:embedding_interp}
\end{equation}

\begin{table}[!ht]
\centering
\scriptsize
\caption{Multiverse embedding-based blending generation results across mixing ratios.}
\label{tab:embedding_blend_proposed}
\begin{tabular}{c c c c c}
\toprule
\textbf{Group} & \textbf{Ratio A} & \textbf{Ratio B} 
& \textbf{ViTScore A} & \textbf{ViTScore B} \\
\midrule

\multirow{5}{*}{\makecell{Intra-genre \\
(G1--G2 \\
G3--G4)}}
& \cellcolor{gray!5}0.00 & \cellcolor{gray!20}1.00 
& \cellcolor{gray!7}0.392{\tiny$\pm$0.005} 
& \cellcolor{gray!18}0.664{\tiny$\pm$0.011} \\

& \cellcolor{gray!10}0.25 & \cellcolor{gray!18}0.75 
& \cellcolor{gray!6}0.379{\tiny$\pm$0.006} 
& \cellcolor{gray!17}0.650{\tiny$\pm$0.015} \\

& \cellcolor{gray!15}0.50 & \cellcolor{gray!15}0.50 
& \cellcolor{gray!13}0.501{\tiny$\pm$0.008} 
& \cellcolor{gray!12}0.493{\tiny$\pm$0.011} \\

& \cellcolor{gray!18}0.75 & \cellcolor{gray!10}0.25 
& \cellcolor{gray!17}0.645{\tiny$\pm$0.012} 
& \cellcolor{gray!7}0.394{\tiny$\pm$0.006} \\

& \cellcolor{gray!20}1.00 & \cellcolor{gray!5}0.00 
& \cellcolor{gray!20}0.671{\tiny$\pm$0.011} 
& \cellcolor{gray!5}0.371{\tiny$\pm$0.004} \\
\midrule

\multirow{5}{*}{\makecell{Inter-genre \\
(G1/G2-- \\
G3/G4)}}
& \cellcolor{gray!5}0.00 & \cellcolor{gray!20}1.00 
& \cellcolor{gray!9}0.411{\tiny$\pm$0.004} 
& \cellcolor{gray!18}0.664{\tiny$\pm$0.010} \\

& \cellcolor{gray!10}0.25 & \cellcolor{gray!18}0.75 
& \cellcolor{gray!10}0.418{\tiny$\pm$0.005} 
& \cellcolor{gray!17}0.647{\tiny$\pm$0.009} \\

& \cellcolor{gray!15}0.50 & \cellcolor{gray!15}0.50 
& \cellcolor{gray!13}0.497{\tiny$\pm$0.009} 
& \cellcolor{gray!13}0.496{\tiny$\pm$0.008} \\

& \cellcolor{gray!18}0.75 & \cellcolor{gray!10}0.25 
& \cellcolor{gray!17}0.643{\tiny$\pm$0.009} 
& \cellcolor{gray!9}0.421{\tiny$\pm$0.004} \\

& \cellcolor{gray!20}1.00 & \cellcolor{gray!5}0.00 
& \cellcolor{gray!19}0.665{\tiny$\pm$0.010} 
& \cellcolor{gray!9}0.423{\tiny$\pm$0.005} \\
\bottomrule
\end{tabular}
\end{table}

\textbf{Interpolation controllability.}
Table~\ref{tab:embedding_blend_proposed} summarizes the blending performance of the proposed model, showing how similarity scores shift across interpolation ratios for each game pair, with aggregated results reported for intra-domain, inter-domain, and overall settings. A controllable blending mechanism should exhibit monotonic behavior, where similarity to an anchor game increases as its interpolation weight becomes larger.
We quantify this tendency using the Pearson correlation~\cite{benesty2009pearson} between the interpolation ratio and similarity scores, obtaining an average correlation of 0.527 across all pairs.
Fig. \ref{fig:embedding_blending} further visualizes per-game blending behavior and shows clear ratio-dependent trends: as the interpolation weight shifts, the generated levels progressively move toward the corresponding anchor domain. Taken together, these results demonstrate that the shared embedding space learned by Multiverse enables smooth, ratio-controllable transitions between domains, providing empirical evidence that the proposed embedding-based approach affords controllable level generation.

\textbf{Training objective ablation.}
Table~\ref{tab:embedding_blend_wgeo} presents an ablation study of training objectives for level blending, comparing a single-positive CL baseline, two ablation variants, and the proposed model. 
Statistical significance is assessed using a paired t-test. 
The table reports controllability via the correlation between interpolation ratio and similarity, as well as generation quality under balanced ($\text{WGM}_{\text{balance}}$, $\alpha=.5$) and biased ($\text{WGM}_{\text{bias}}$, $\alpha=.75$) blending settings.

\begin{table}[!ht]
\centering
\caption{
Comparison of generation models for level blending. Correlation evaluates controllability of embedding interpolation. Generation quality is measured using weighted geometric mean scores.
}
\label{tab:embedding_blend_wgeo}
\setlength{\tabcolsep}{3pt}
\begin{tabular}{llccc}
\toprule
\textbf{Group} & \textbf{Method} 
& \textbf{Correlation} 
& \textbf{$\boldsymbol{\mathrm{WGM}_{\text{balance}}}$} 
& \textbf{$\boldsymbol{\mathrm{WGM}_{\text{bias}}}$} \\
\midrule

\multirow{3}{*}{Intra-genre}
& Single CL      & 0.558{\scriptsize$\pm$0.017} & 0.477{\scriptsize$\pm$0.004} & 0.549{\scriptsize$\pm$0.006} \\
& -Abstraction  & \down{0.548{\scriptsize$\pm$0.016}} & \down{0.473{\scriptsize$\pm$0.004}}* & \down{0.546{\scriptsize$\pm$0.003}} \\
& -Threshold  & \down{0.556{\scriptsize$\pm$0.012}} & \up{0.479{\scriptsize$\pm$0.004}} & \down{0.548{\scriptsize$\pm$0.006}} \\
& \skycell Multiverse  & \skycell \textbf{\up{0.580{\scriptsize$\pm$0.019}}*} & \skycell \textbf{\up{0.481{\scriptsize$\pm$0.009}}} & \skycell \textbf{\up{0.559{\scriptsize$\pm$0.008}}*} \\
\midrule

\multirow{3}{*}{Inter-genre}
& Single CL      & 0.482{\scriptsize$\pm$0.012} & \textbf{0.493{\scriptsize$\pm$0.008}} & 0.563{\scriptsize$\pm$0.004} \\
& -Abstraction  & \down{0.469{\scriptsize$\pm$0.012}}* & \down{0.492{\scriptsize$\pm$0.007}} & \down{0.558{\scriptsize$\pm$0.004}}* \\
& -Threshold  & \down{0.479{\scriptsize$\pm$0.008}} & \textbf{0.493{\scriptsize$\pm$0.008}} & \down{0.561{\scriptsize$\pm$0.005}} \\
& \skycell Multiverse  & \skycell \textbf{\up{0.498{\scriptsize$\pm$0.020}}} & \skycell \down{0.479{\scriptsize$\pm$0.008}}** & \skycell \textbf{\up{0.568{\scriptsize$\pm$0.007}}} \\
\midrule

\multirow{3}{*}{Overall}
& Single CL      & 0.509{\scriptsize$\pm$0.013} & \textbf{0.488{\scriptsize$\pm$0.006}} & 0.558{\scriptsize$\pm$0.004} \\
& -Abstraction  & \down{0.497{\scriptsize$\pm$0.012}}* & \down{0.485{\scriptsize$\pm$0.005}} & \down{0.553{\scriptsize$\pm$0.003}}* \\
& -Threshold  & \down{0.506{\scriptsize$\pm$0.007}} & \textbf{0.488{\scriptsize$\pm$0.006}} & \down{0.556{\scriptsize$\pm$0.004}} \\
& \skycell Multiverse & \skycell \textbf{\up{0.527{\scriptsize$\pm$0.019}}} & \skycell \down{0.480{\scriptsize$\pm$0.007}} & \skycell \textbf{\up{0.565{\scriptsize$\pm$0.006}}*} \\
\bottomrule
\end{tabular}

\vspace{2pt}
{\footnotesize \hfill Significance vs.\ Single CL: * $p<0.05$, ** $p<0.01$}

\end{table}

Multiverse achieves the highest correlation overall, with a statistically significant improvement in the intra-genre setting ($p=0.034$), indicating improved embedding-level controllability when interpolating between semantically related domains. 
For generation quality, Multiverse shows stronger performance under biased blending, achieving higher $\text{WGM}_{\text{bias}}$ across settings with significant improvements in the intra-genre case ($p=0.013$). 
In contrast, under balanced blending, $\text{WGM}_{\text{balance}}$ remains comparable to the single CL baseline and decreases in the inter-genre case, reflecting the stricter requirement of simultaneously matching both domains.

The ablation results further highlight the importance of the proposed components. 
Removing meta-instruction abstraction (\textit{-Abstraction}) or disabling similarity-based multi-positive supervision (\textit{-Threshold}) generally reduces controllability and blending quality compared to the full model. 
Overall, these results suggest that meta-instruction multi-positive alignment helps stabilize cross-game conditional features, enabling more controllable embedding interpolation while maintaining strong performance under biased blending conditions.

\begin{figure*}[!t]
\centering
\label{fig:level_blending_compact}

\setlength{\tabcolsep}{4pt}  
\renewcommand{\arraystretch}{1}

\begin{tabular}{>{\centering\arraybackslash}p{2.2cm}
                >{\centering\arraybackslash}p{2.2cm}
                >{\centering\arraybackslash}p{2.2cm}
                >{\centering\arraybackslash}p{2.2cm}
                >{\centering\arraybackslash}p{2.2cm}
                >{\centering\arraybackslash}p{2.2cm}}
\toprule
Text A & Text B & Concat & Mix & A-base & B-base \\
\midrule
\includegraphics[width=2.2cm]{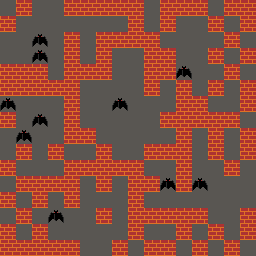} &
\includegraphics[width=2.2cm]{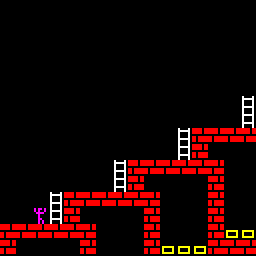} &
\includegraphics[width=2.2cm]{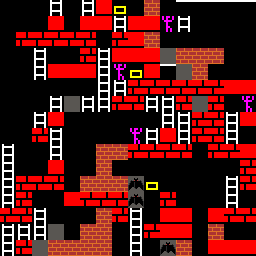} &
\includegraphics[width=2.2cm]{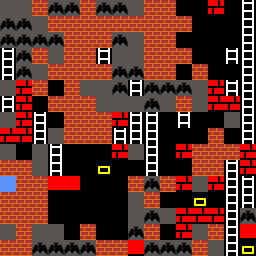} &
\includegraphics[width=2.2cm]{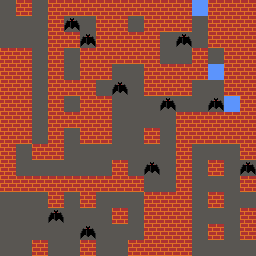}&
\includegraphics[width=2.2cm]{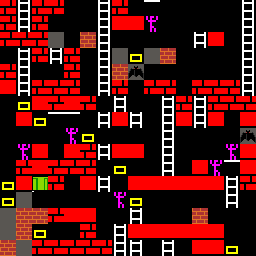} \\
\bottomrule 
\addlinespace[1pt]
\multicolumn{6}{c}{\footnotesize{
\parbox{14.4cm}{\footnotesize
A: Central cross wall with bats scattered around and open path pockets \\
B: Ground bottom band diggable ground central block ladders left center top rope none gold none
}
}
}
\end{tabular}
\caption{\textbf{Multi-game Text Instruction-based Level Blending.}
Levels generated by composite
text instruction between the embeddings of Super Mario (A) and Lode Runner (B) across different instruction combination strategies.}
\vspace{-0.4cm}
\label{fig:text_blending}
\end{figure*}

\subsection{Level Blending via Multi-game Text Instruction}
This experiment investigates whether the generator can produce zero-shot blended levels when conditioned on a composite text instruction that combines two game domains within a single prompt. In this setting, the composite instructions are unseen during training and often exhibit increased linguistic complexity due to the combination of multiple domain-specific descriptions. Consequently, the model must determine which conditions in each sentence are most influential for generation and balance structural cues originating from the different game domains.

Composite instructions are constructed by sampling pairs of game levels from the dataset and combining their corresponding instructions into a single prompt. We consider four instruction combination strategies: direct concatenation of two instructions (\textit{Concat}) and semantic integration of two instructions (\textit{Mix}), which represent balanced combinations; and using the instruction from game A as the base while incorporating conditions from game B (\textit{A-base}), and vice versa (\textit{B-base}), which represent biased combinations. The resulting composite instructions contain $28.96 \pm 9.43$ words on average, approximately twice as long as instructions in the single-instruction dataset.
The detailed procedure for constructing these composite instructions is described in~\cref{sec:multigame_instruction}. 
The generated levels reflect the intended blending ratios defined by the composite instructions, as shown in Fig.~\ref{fig:text_blending}

\begin{figure}[!t]
    \centering
    \includegraphics[width=0.8\linewidth]{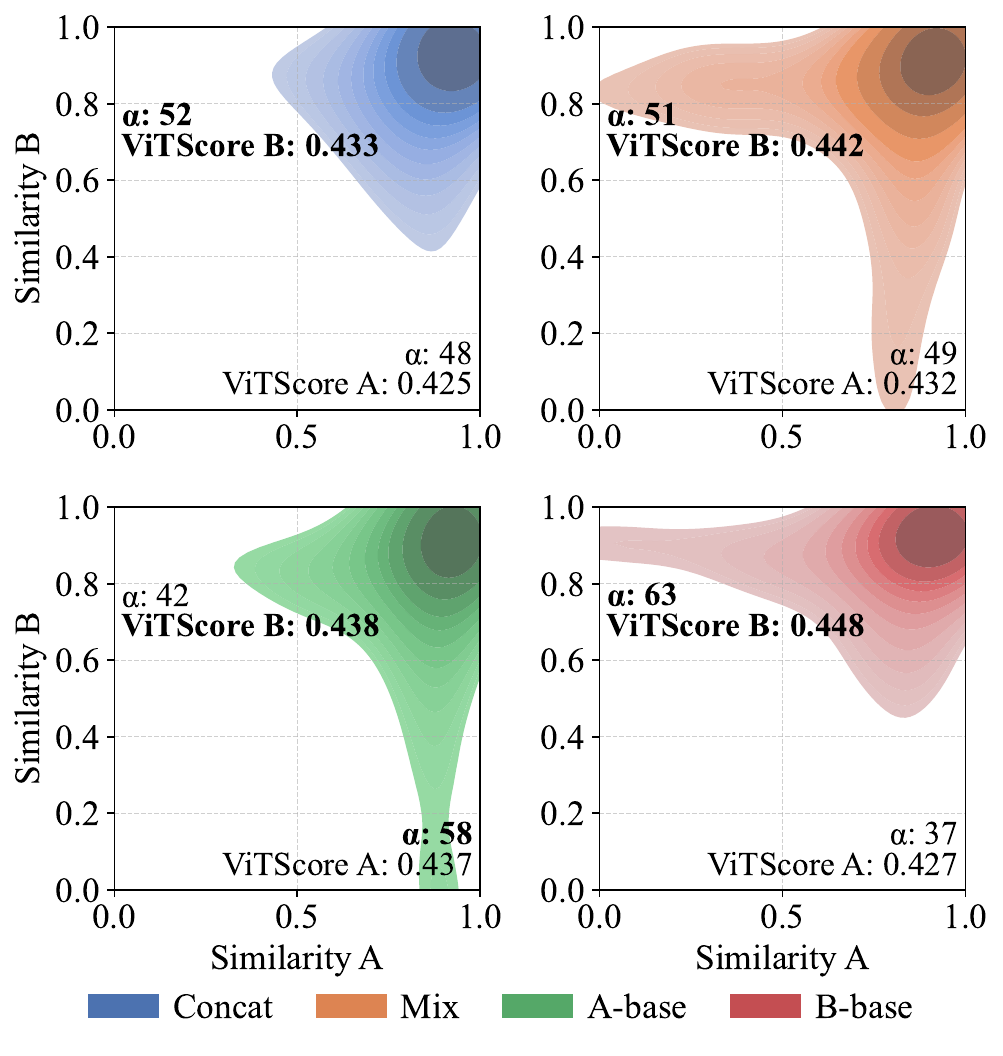}
    \vspace{-0.4cm}
    \caption{\textbf{Cosine Similarity Distribution of Composite Instructions.}
    2D kernel density estimation visualization of cosine similarity between composite instructions and the original instructions from game A and game B. Balanced combinations remain relatively symmetric, while biased combinations shift toward the axis corresponding to the base instruction.
    }
    \label{fig:multi-game_text_instructions}
    \vspace{-0.5cm}
\end{figure}

Fig.~\ref{fig:multi-game_text_instructions} illustrates how these strategies position composite instructions relative to the original game instructions in the embedding space. These results indicate how instruction composition strategies influence the semantic characteristics of composite prompts. Balanced combinations produce embeddings that remain relatively equidistant from both source instructions, whereas biased combinations shift the embedding toward the base domain, reflecting stronger semantic alignment with that domain. This behavior can be interpreted as implicit interpolation in the embedding space. The interpolation ratios further reveal the interpolation patterns of each strategy.
The interpolation ratio is computed from the $\alpha$ value in Eq.~\eqref{eq:embedding_interp}. Among the balanced strategies, \textit{Concat} produces an interpolation of 48:52 and \textit{Mix} produces 49:51, indicating nearly symmetric contributions from both domains; in contrast, the biased strategies shift the ratio toward one domain, with \textit{A-base} yielding 58:42 and \textit{B-base} yielding 37:63.
Although \textit{Concat} and \textit{Mix} produce nearly symmetric interpolation ratios, their distributions differ significantly. \textit{Concat} results in a more compact and stable embedding, whereas \textit{Mix} leads to a broader and more dispersed distribution, suggesting that even with similar interpolation ratios, the distribution of embeddings varies depending on how the instructions are composed.
These interpolation ratios and mirrored patterns indicate that instruction composition strategies systematically shape the interpolation behavior of composite instructions.
These results show that Multiverse enables cross-game level blending using composite instructions, where the relative contribution of each domain can be controlled by the structure of the instruction.


\section{Conclusion and Future Work}
This paper presents Multiverse, a language-conditioned multi-game level generator that unifies multi-game generation and cross-game level blending within a shared latent space. 
The model jointly aligns text and level representations while explicitly connecting heterogeneous game domains through a cross-domain connectivity learning objective. 
Experimental results show that the learned representation produces ratio-consistent blending behavior and improves blending controllability compared to a contrastive learning baseline. 
By integrating multiple game domains within a shared representation space, the proposed approach enables a unified generative model across games and introduces cross-domain level blending as a new mechanism for hybrid level design.

One limitation of the proposed approach is the relatively weak bias observed in text-based interpolation. Although biased composite instructions are intended to emphasize one domain over the other, the generated levels do not always exhibit a strongly biased transition toward the designated domain. 
This likely stems from the current training setup, which is trained on single-domain instructions and therefore does not explicitly learn combinational instruction structures, despite its potential for multi-game instruction learning. Future work will explore training strategies that explicitly model diverse compositional instructions and mixing intents to enable more controllable text-driven level blending.

\bibliographystyle{IEEEtran}
\bibliography{references}

@inproceedings{khalifa2020pcgrl,
  title={Pcgrl: Procedural content generation via reinforcement learning},
  author={Khalifa, Ahmed and Bontrager, Philip and Earle, Sam and Togelius, Julian},
  booktitle={Proceedings of the AAAI Conference on Artificial Intelligence and Interactive Digital Entertainment},
  volume={16},
  number={1},
  pages={95--101},
  year={2020}
}

@inproceedings{merino2023five,
  title={The five-dollar model: generating game maps and sprites from sentence embeddings},
  author={Merino, Timothy and Negri, Roman and Rajesh, Dipika and Charity, M and Togelius, Julian},
  booktitle={Proceedings of the AAAI Conference on Artificial Intelligence and Interactive Digital Entertainment},
  volume={19},
  number={1},
  pages={107--115},
  year={2023}
}

@inproceedings{fontaine2021illuminating,
  title={Illuminating mario scenes in the latent space of a generative adversarial network},
  author={Fontaine, Matthew C and Liu, Ruilin and Khalifa, Ahmed and Modi, Jignesh and Togelius, Julian and Hoover, Amy K and Nikolaidis, Stefanos},
  booktitle={Proceedings of the AAAI Conference on Artificial Intelligence},
  volume={35},
  number={7},
  pages={5922--5930},
  year={2021}
}

@inproceedings{radford2021learning,
  title={Learning transferable visual models from natural language supervision},
  author={Radford, Alec and Kim, Jong Wook and Hallacy, Chris and Ramesh, Aditya and Goh, Gabriel and Agarwal, Sandhini and Sastry, Girish and Askell, Amanda and Mishkin, Pamela and Clark, Jack and others},
  booktitle={International conference on machine learning},
  pages={8748--8763},
  year={2021},
  organization={PmLR}
}

@inproceedings{baek2025ipcgrl,
  title={IPCGRL: Language-Instructed Reinforcement Learning for Procedural Level Generation},
  author={Baek, In-Chang and Kim, Sung-Hyun and Lee, Seo-Young and Kim, Dong-Hyeon and Kim, Kyung-Joong},
  booktitle={2025 IEEE Conference on Games (CoG)},
  pages={1--8},
  year={2025},
  organization={IEEE}
}

@article{abdullah2024chatgpt4pcg,
  title={The 1 st ChatGPT4PCG Competition},
  author={Abdullah, Febri and Taveekitworachai, Pittawat and Dewantoro, Mury F and Thawonmas, Ruck and Togelius, Julian and Renz, Jochen},
  journal={IEEE Transactions on Games},
  year={2024},
  publisher={IEEE}
}

@article{atmakuri2025semi,
  title={Semi-Supervised Tile Embeddings: A General, Multi-Game Level Representation},
  author={Atmakuri, Venkata Sai Revanth and Satvati, Kian Razavi and Sarkar, Anurag and Guzdial, Matthew},
  journal={IEEE Transactions on Games},
  year={2025},
  publisher={IEEE}
}

@article{sarkar2020controllable,
  title={Controllable level blending between games using variational autoencoders},
  author={Sarkar, Anurag and Yang, Zhihan and Cooper, Seth},
  journal={arXiv preprint arXiv:2002.11869},
  year={2020}
}

@inproceedings{kumaran2019generating,
  title={Generating game levels for multiple distinct games with a common latent space},
  author={Kumaran, Vikram and Mott, Bradford and Lester, James},
  booktitle={Proceedings of the AAAI conference on artificial intelligence and interactive digital entertainment},
  volume={15},
  number={1},
  pages={102--108},
  year={2019}
}

@article{sudhakaran2023mariogpt,
  title={Mariogpt: Open-ended text2level generation through large language models},
  author={Sudhakaran, Shyam and Gonz{\'a}lez-Duque, Miguel and Freiberger, Matthias and Glanois, Claire and Najarro, Elias and Risi, Sebastian},
  journal={Advances in Neural Information Processing Systems},
  volume={36},
  pages={54213--54227},
  year={2023}
}

@article{sarkar2023procedural,
  title={Procedural content generation via knowledge transformation (PCG-KT)},
  author={Sarkar, Anurag and Guzdial, Matthew and Snodgrass, Sam and Summerville, Adam and Machado, Tiago and Smith, Gillian},
  journal={IEEE Transactions on Games},
  volume={16},
  number={1},
  pages={36--50},
  year={2023},
  publisher={IEEE}
}

@article{VGLC,
Author = {Adam James Summerville and Sam Snodgrass and Michael Mateas and Santiago Ontañón},

Title = {The VGLC: The Video Game Level Corpus},

Year = {2016},

Journal = {Proceedings of the 7th Workshop on Procedural Content Generation},

}

@article{van2017neural,
  title={Neural discrete representation learning},
  author={Van Den Oord, Aaron and Vinyals, Oriol and others},
  journal={Advances in neural information processing systems},
  volume={30},
  year={2017}
}

@inproceedings{kaiser2018fast,
  title={Fast decoding in sequence models using discrete latent variables},
  author={Kaiser, Lukasz and Bengio, Samy and Roy, Aurko and Vaswani, Ashish and Parmar, Niki and Uszkoreit, Jakob and Shazeer, Noam},
  booktitle={International Conference on Machine Learning},
  pages={2390--2399},
  year={2018},
  organization={PMLR}
}

@article{baek2025human,
  title={Human-Aligned Procedural Level Generation Reinforcement Learning via Text-Level-Sketch Shared Representation},
  author={Baek, In-Chang and Lee, Seoyoung and Kim, Sung-Hyun and Hwang, Geumhwan and Kim, KyungJoong},
  journal={arXiv preprint arXiv:2508.09860},
  year={2025}
}

@article{dosovitskiy2020vit,
  title={An Image is Worth $16\times16$ Words: Transformers for Image Recognition at Scale},
  author={Dosovitskiy, Alexey and Beyer, Lucas and Kolesnikov, Alexander and Weissenborn, Dirk and Zhai, Xiaohua and Unterthiner, Thomas and  Dehghani, Mostafa and Minderer, Matthias and Heigold, Georg and Gelly, Sylvain and Uszkoreit, Jakob and Houlsby, Neil},
  journal={ICLR},
  year={2021}
}

@article{kim2025multi,
  title={Multi-Objective Instruction-Aware Representation Learning in Procedural Content Generation RL},
  author={Kim, Sung-Hyun and Baek, In-Chang and Lee, Seo-Young and Hwang, Geum-Hwan and Kim, Kyung-Joong},
  journal={arXiv preprint arXiv:2508.09193},
  year={2025}
}

@article{zhu2024evaluate,
  title={How to evaluate semantic communications for images with vitscore metric?},
  author={Zhu, Tingting and Peng, Bo and Liang, Jifan and Han, Tingchen and Wan, Hai and Fu, Jingqiao and Chen, Junjie},
  journal={IEEE Transactions on Cognitive Communications and Networking},
  volume={10},
  number={5},
  pages={1744--1758},
  year={2024},
  publisher={IEEE}
}

@incollection{benesty2009pearson,
  title={Pearson correlation coefficient},
  author={Benesty, Jacob and Chen, Jingdong and Huang, Yiteng and Cohen, Israel},
  booktitle={Noise reduction in speech processing},
  pages={1--4},
  year={2009},
  publisher={Springer}
}

@inproceedings{sarkar2021generating,
  title={Generating and blending game levels via quality-diversity in the latent space of a variational autoencoder},
  author={{A. Sarkar and S. Cooper}},
  booktitle={Proceedings of the 16th International Conference on the Foundations of Digital Games},
  pages={1--11},
  year={2021}
}

@inproceedings{nie2025moonshine,
  title={Moonshine: Distilling game content generators into steerable generative models},
  author={Nie, Yuhe and Middleton, Michael and Merino, Tim and Kanagaraja, Nidhushan and Kumar, Ashutosh and Zhuang, Zhan and Togelius, Julian},
  booktitle={Proceedings of the AAAI Conference on Artificial Intelligence},
  volume={39},
  number={13},
  pages={14344--14351},
  year={2025}
}

@inproceedings{sarkar2021dungeon,
  title={Dungeon and platformer level blending and generation using conditional vaes},
  author={Sarkar, Anurag and Cooper, Seth},
  booktitle={2021 IEEE Conference on Games (CoG)},
  pages={1--8},
  year={2021},
  organization={IEEE}
}

\appendix

\subsection{Multi-game Text Instruction}
\label{sec:multigame_instruction}
Multi-game text instructions serve as conditioning prompts, where a single instruction combines two game domains to evaluate whether the generator produces blended levels. Let $T_A$ denote the instruction from game A and $T_B$ from game B. Four instruction combination strategies represent different levels of semantic integration. 
Example composite instructions for each strategy are shown in Table~\ref{tab:composite_instruction}.

\textbf{Concat} concatenates $T_A$ and $T_B$ sequentially without modification. \textbf{Mix} produces a semantically integrated instruction that merges the contents of $T_A$ and $T_B$ into a single coherent description. Biased instructions follow the template \textit{based on $T_{\text{base}}$, featuring $T_{\text{aux}}$}. In \textbf{A-base}, $T_A$ provides the primary structure while elements from $T_B$ appear as auxiliary features; \textbf{B-base} reverses these roles. The composite instructions for Mix, A-base, and B-base are generated using \texttt{gpt-5.2}. The model receives the two source instructions ($T_A$ and $T_B$) together with a written guideline specifying how the instructions should be combined for each strategy (Mix, A-base, and B-base), ensuring that the generated prompts follow the intended composition pattern.

\begin{table}[!h]
\caption{\textbf{Example of composite instruction construction.}
Text from Game A (\textit{The Legend of Zelda}) and Game B (\textit{Dungeon}) are highlighted.}
\centering
\begin{tabular}{p{0.8cm} p{7.0cm}}
\toprule
\textbf{Strategy} & \textbf{Composite Instruction} \\
\midrule

Text A &
\hlA{square floor room} with \hlA{central block cluster} and \hlA{surrounding wall} \\

Text B &
\hlB{spiral wall path} with \hlB{bats clustered on right side} \\

\midrule

Concat &
\hlA{square floor room} with \hlA{central block cluster} and \hlA{surrounding wall} \hlB{spiral wall path} with \hlB{bats clustered on right side} \\

Mix &
\hlA{A square-floored room} is enclosed by a continuous wall that \hlB{folds into a spiral path} around a \hlA{central cluster} \hlA{of blocks}, and along the \hlB{right side of the winding wall} \hlB{the bats gather in a dense, shadowy group}. \\

A-base &
\textbf{based on} \hlA{square floor room with central block cluster and} \hlA{surrounding wall}, \textbf{featuring} \hlB{spiral wall path with bats clustered} \hlB{on right side}. \\

B-base &
\textbf{based on} \hlB{spiral wall path with bats clustered on right side}, \textbf{featuring} a \hlA{square floor room with a central block cluster and} \hlA{surrounding wall}. \\

\bottomrule
\end{tabular}
\label{tab:composite_instruction}
\end{table}

\end{document}